\documentclass{article}

\usepackage[nonatbib,final]{nips_2016}

\usepackage[utf8]{inputenc} 
\usepackage[T1]{fontenc}    
\usepackage{hyperref}       
\usepackage{url}            
\usepackage{booktabs}       
\usepackage{amsfonts}       
\usepackage{nicefrac}       
\usepackage{microtype}      
\usepackage{titlesec}

\usepackage[colorinlistoftodos]{todonotes}

\title{SocialML: machine learning\\ for social media video creators}

\author{
  Tomasz Trzcinski$^{a,b}$, Adam Bielski$^{b}$, Pawel Cyrta$^{b}$ and Matthew Zak$^{b}$\\
  $^{a}$Warsaw University of Technology $^{b}$ Tooploox\\
  \texttt{firstname.lastname@tooploox.com}
}

\begin{document}

\titlespacing*{\maketitle}{0pt}{1ex plus 0.5ex minus .1ex}{0.5ex plus .1ex}
\titlespacing*{\section}{0pt}{0.0ex plus 0.0ex minus 0.6ex}{0.0ex plus .0ex minus 0.3ex}
\titlespacing*{\paragraph}{0pt}{-0.2ex plus 0.2ex minus 0.5ex}{0.4ex plus .0ex}

\maketitle

\section*{Abstract}

In the recent years, social media have become one of the main places where creative content is being published and consumed by billions of users. 

Contrary to traditional media, social media allow the publishers to receive almost instantaneous feedback regarding their creative work at an unprecedented scale. This is a perfect use case for machine learning methods that can use these massive amounts of data to provide content creators with inspirational ideas and constructive criticism of their work.  

In this work, we present a comprehensive overview of machine learning-empowered tools we developed for video creators at Group Nine Media - one of the major social media companies that creates short-form videos with over three billion views per month. 

Our main contribution is a set of tools that allow the creators to leverage massive amounts of data to improve their creation process, evaluate their videos before the publication and improve content quality. These applications include an interactive conversational bot that allows access to material archives, a Web-based application for automatic selection of optimal video thumbnail, as well as deep learning methods for optimizing headline and predicting video popularity. Our A/B tests show that deployment of our tools leads to significant increase of average video view count by 12.9\%. Our additional contribution is a set of considerations collected during the deployment of those tools that can help to understand the challenges of applying machine learning methods in creative practice.

\section*{Tools} 

\paragraph{Assisted thumbnail selection.}

A video thumbnail is often the first information about the video a social media user can see. It is therefore critical to choose it wisely to maximize the chances of catching viewer's attention. Typically, the selection of the thumbnail is done based on publisher's intuition and experience. Using the consumption data of social media videos from the past, we built a machine learning algorithm to select the thumbnail for social media video that is likely to attract users' attention and recommend it to the publisher.

\begin{figure}[h]
  \centering
  \includegraphics[scale=0.5]{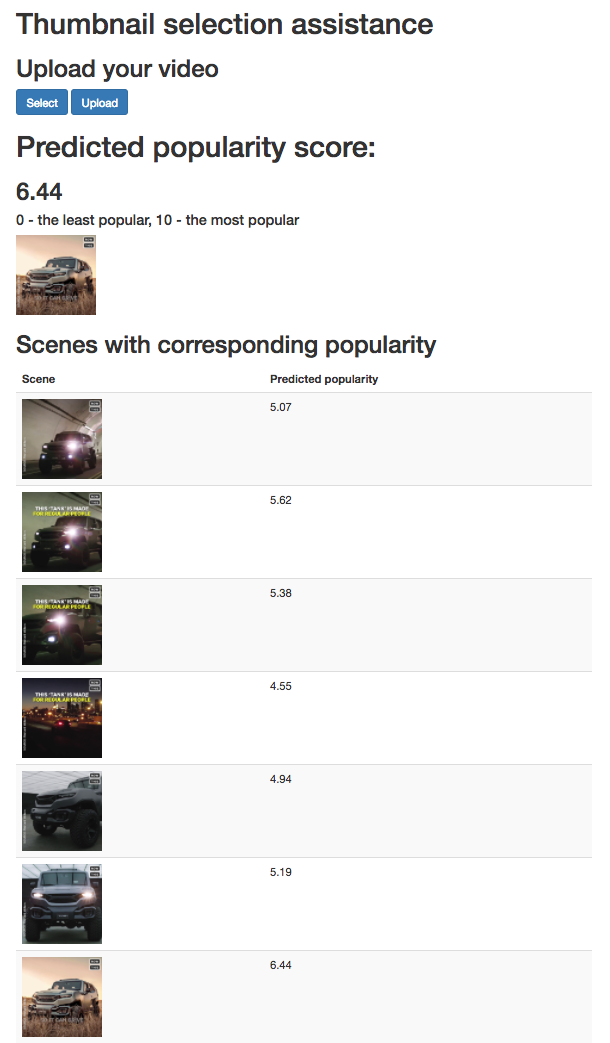}
  \caption{Web application for thumbnail selection assistance. Subsets of analyzed frames with corresponding scores are shown. Our model uses fine-tuned convolutional neural network to analyse a set of uniformly sampled frames from a video and outputs a corresponding popularity score. Recommended thumbnail is selected as the frame with the highest score returned.}
  \label{fig:thumbnail-selection}
\end{figure}

For the purpose of our training, we collected a dataset of 37’042 Facebook videos along with their thumbnails and titles from top publishers according to TubularLabs.com ranking. To take into account the influence of channel popularity on individual video popularity, we normalized number of views of each video by dividing it by the number of likes of a channel that the video was posted on. We then split the dataset into popular and unpopular class using median normalized view count. We trained a model for binary classification by fine-tuning the last layer of ResNet50 \cite{resnet} model pre-trained on ImageNet dataset \cite{imagenet}. We used 80\% of data for training, 10\% for validation and 10\% for testing. We reached 66\% classification accuracy on the test dataset. In production, we apply the model to 40 frames extracted uniformly from a video. The frame evaluated with the highest score is recommended as a thumbnail. A screenshot of the application can be seen in Fig.~\ref{fig:thumbnail-selection}.

\paragraph{Slack chatbot for researching archived materials.}

Quick and intuitive access to materials is a prerequisite for effective work of any video creator. In order to facilitate it, we implemented a responsive chatbot that returns lists of videos tagged with topics related to user's query. Materials are indexed using meta-data extracted from videos using image and text processing. More precisely, we use TextBoxes \cite{TextBoxes} 

to detect subtitles and other relevant text snippets in the videos, and then we recognize the text using CRNN \cite{crnn}. Then we input video headline and text extracted from the video into a neural network topic classifier consisting of dense layer, ReLU and a softmax layer for two tier category classes. As text embedding, we use 400-dimensional FastText vector representation~\cite{bojanowski2016enriching}. Similar tagging is applied to audio transcription generated using speakers diarization segmentation ~\cite{cyrta2017speaker} and speech recognition models.

\begin{figure}[h]
  \centering
  \includegraphics[scale=0.45]{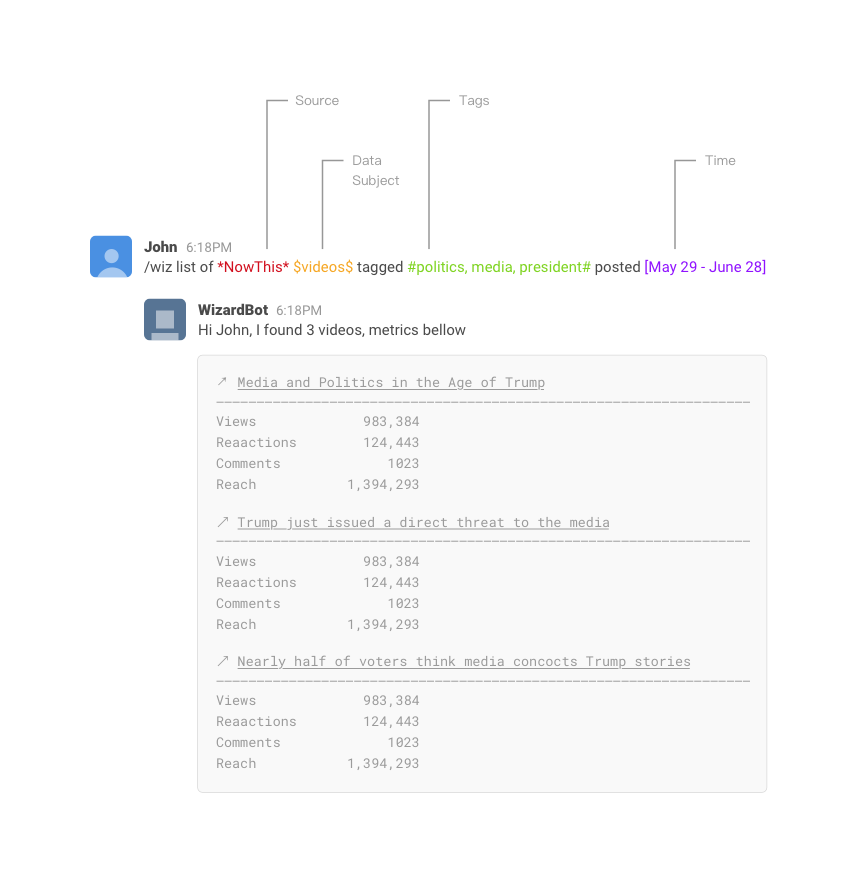}
  \caption{Example of a query question posed at the Slack bot for indexing archive materials. As a result, the bot returns social media interaction statistics related to all videos with similar tags. The tagging is done offline using textual input from both headline and text displayed in the video frames.}
  \label{fig:wizardbot-query}
\end{figure}

\begin{figure}[h]
  \centering
  \includegraphics[scale=0.4]{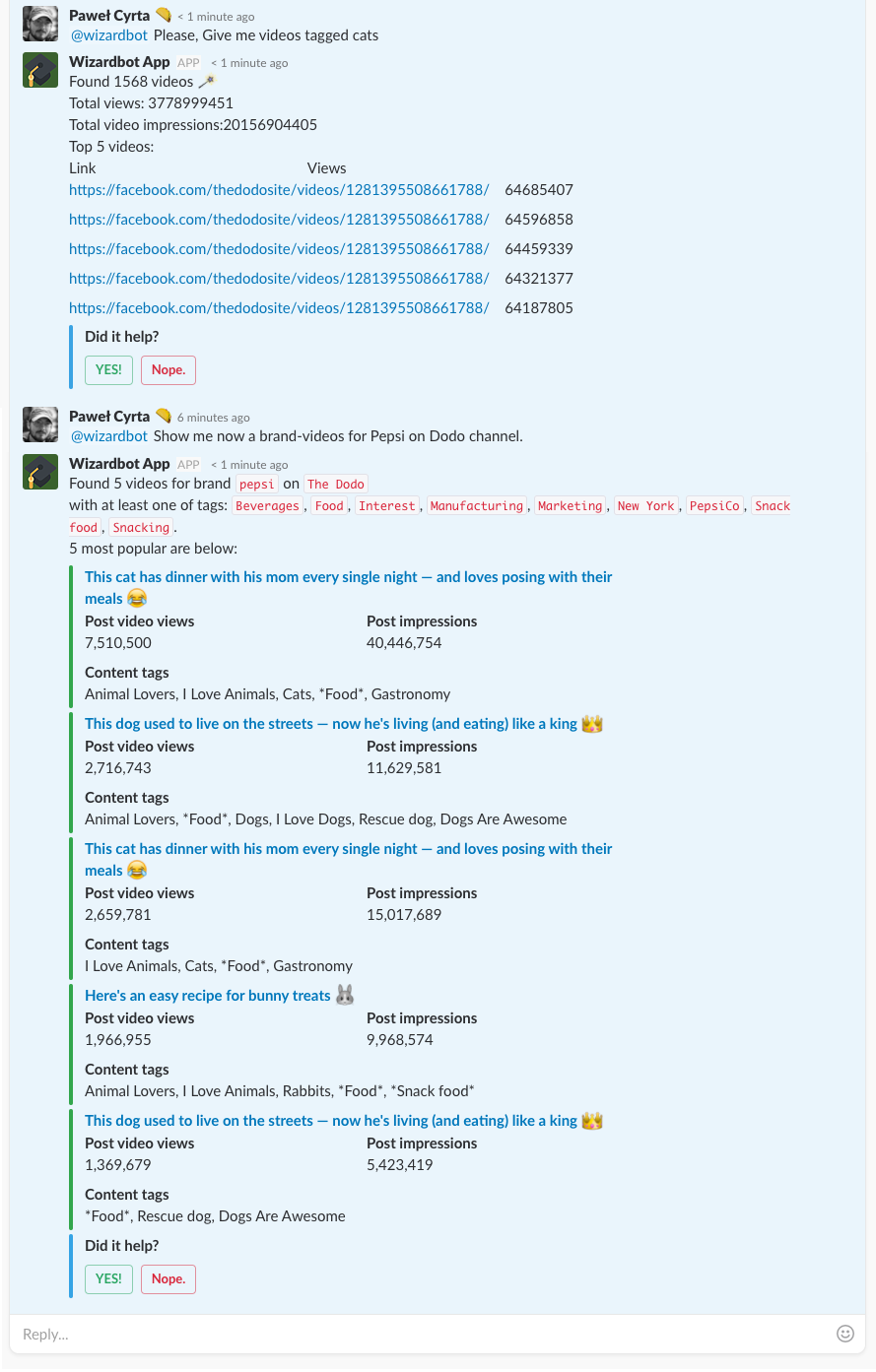}
  \caption{Sample conversation with a Slack bot using natural language understanding module. The bot is able to retrieve various statistics related to material published in the past.}
  \label{fig:wizardbot-dialogue}
\end{figure}

After interviewing a sample of our users, we decided to deploy the service using Slack. User can ask our Slack bot several questions related to retrieving indexed content, e.g. related to user interaction statistics of videos with a given tag. We used standard decision tree rule-based and slot matching methods for the natural language understanding component of our bot. A screenshot of a sample conversation with Slack bot can be seen in Fig.~\ref{fig:wizardbot-query} and Fig. ~\ref{fig:wizardbot-dialogue}.

\paragraph{Headline optimization.}
Video titles are short video descriptions displayed above the content and written to attract users' attention. To improve the effectiveness of those headlines, we trained a deep learning recurrent model that scores proposed title through indirect modeling of users' preferences. 

\begin{figure}[h]
  \centering
  \includegraphics[scale=0.6]{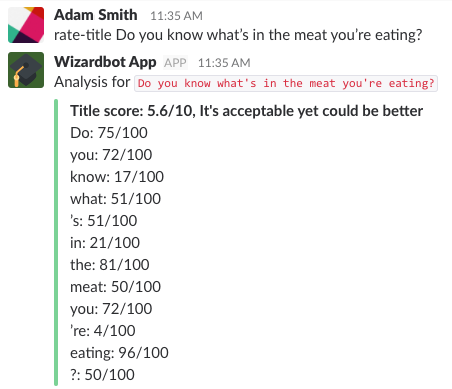}
  \caption{Sample use case of headline optimization model API deployed as a Slack bot functionality. Using bi-directional LSTM neural network model with attention, we are able to estimate popularity of a video as well as determine contributions of individual words.}
  \label{fig:ratemytitle}
\end{figure}

We used the dataset of Facebook videos described above along with their normalized view counts. Inspired by~\cite{Stokowiec17}, we implemented a bi-directional LSTM network with attention \cite{NMTattention} and trained it with video titles transformed with pre-trained GloVe embeddings~\cite{glove}. The objective of our network was popularity classification and the model achieved 70\% accuracy on the test dataset. Thanks to the attention mechanism we can visualize how specific words contribute to popularity of the video and guide publisher's creative process of improving it. We deployed the tool in production through the Slack bot described above. A screenshot of this functionality can be seen in Fig.~\ref{fig:ratemytitle}.

\paragraph{Video popularity prediction based on its frames.}
Incrementing a video view count after first few seconds and setting Facebook videos to auto-play by default leads to increased importance of the first few seconds of the video published on this platform. We therefore leverage past consumption data to help publishers improve the quality of the opening scene and hence lead to higher video popularity and viewer retention. To that end, we implemented a deep neural network architecture that analyses first frames of a video and predicts its future popularity.

\begin{figure}[h]
  \centering
  \includegraphics[scale=0.75]{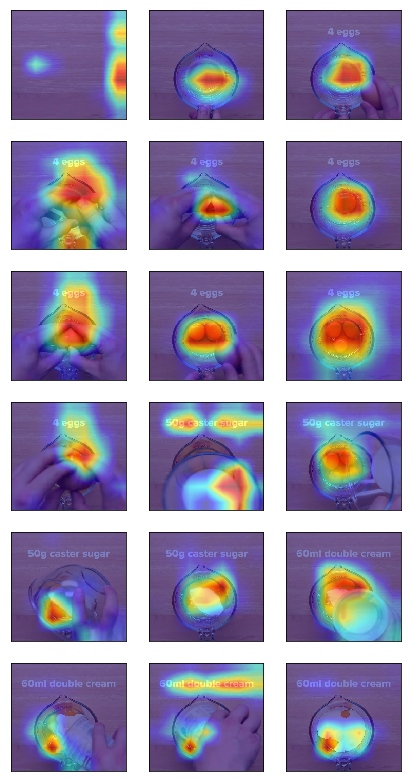}
  \caption{Sample GradCAM visualizations of a video opening scene frames generated using deep neural network trained for popularity classification. The visualizations can be used by video editors to analyze the importance of specific frame parts to social media users.}
  \label{fig:gradcam-sample}
\end{figure}

\begin{figure}[h]
  \centering
  \includegraphics[scale=0.75]{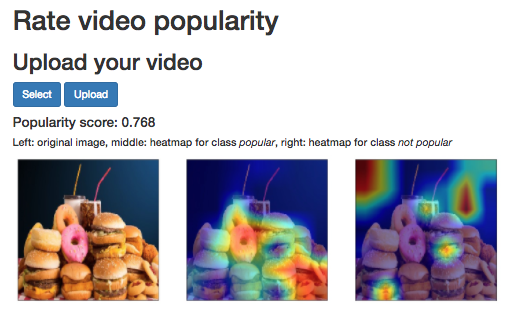}
  \caption{A working Web-based application for predicting future popularity of a video using deep visual features. Our model uses deep visual features extracted from penultimate layer of the ResNet50 model to classify content as popular or unpopular. The probability of class 'popular' is used as popularity score displayed in the application. Furthermore, GradCAM visualisation is used to identify parts of the frame contributing to the popular (middle image) and unpopular (right image) class.}
  \label{fig:gradcam-demo}
\end{figure}

To train our model, we used the dataset of Facebook videos described above and extracted 18 evenly distributed frames from the first 6 seconds of a video. For each frame, we extract features using penultimate layer of the ResNet50~\cite{resnet} model pre-trained on ImageNet~\cite{imagenet} dataset. We follow~\cite{Trzcinski17} and train a neural network model with attention mechanism to classify video as popular or not. The resulting model achieved over 68\% classification accuracy on the test dataset. Inspired by~\cite{Bylinskii17}, we wanted to visualize what parts of the image contribute to popularity score. Hence, we used GradCAM \cite{gradcam} to create heatmaps overlaid on the original frames. This way we can identify parts of frames that contribute the most to video popularity, as in Fig.~\ref{fig:gradcam-sample}. We deployed the tool as a Web service. Fig.~\ref{fig:gradcam-demo} shows a~screenshot of a working application along with a popularity heatmap.

\section*{Applying machine learning in creative practices}

When developing our tools, 
we faced several difficulties related to socio-cultural aspects of using machine learning in creative practice, which can be summarized in the following questions:

\paragraph{How to deploy machine learning algorithms to assess content quality without the risk of offending its creators?} 

Journalists often consider creativity as their most valuable talent and they are reluctant to allow machines to quantify it. We therefore introduced our tools for popularity prediction as an automatic system to alert the editor if their video was expected to be particularly unpopular. In practice, the deployed implementation analyses every video uploaded through a Web user interface and if the normalized popularity score falls below 20-th percentile score, the user is alerted through a pop up window. This significantly increased the acceptance rate of the tool within our user base.

\paragraph{How to validate the impact of a machine learning system in creative practice?} 

To better understand the impact of our toolkit on the work of video creators, we ran A/B tests on user groups working with and without our tools. We then analyzed content popularity metrics, such as video views, shares and comments, for the videos created by both groups. 

Although using these metrics as content quality estimators has several shortcomings (such as sensitiveness to clickbaity news titles), they remain main evaluation metrics used by creators themselves and, hence, we adopted them too.

\paragraph{How to avoid the situation where all the creative content converges on one topic that is intrinsically linked to higher popularity? }

This bias of a popularity estimation algorithm, related closely to the bias present in the training dataset, also exists in other machine learning applications~\cite{Tommasi15}. The solution we proposed to avoid it in our application is to categorize the content and normalize its popularity scores according to the median category value. By using this additional normalization step, our method is more likely to concentrate on the features related to content quality instead of its topic.

\section*{Acknowledgment}
The authors would like to thank Group Nine Media Inc. for enabling this research.

\small

\bibliographystyle{ieeetr}
\bibliography{sample.bib}

\clearpage

\section*{Figures}

\end{document}